# Dual-discriminator GAN: A GAN way of profile face recognition

XINYU ZHANG Department of Electronic Engineering Tsinghua University Beijing, 100084, P.R. China (e-mail: xy-zhang17@mails.tsinghua.edu.cn)

YANG ZHAO Department of Electronic Engineering Tsinghua University Beijing, 100084, P.R. China (e-mail: zhao-yan18@mails.tsinghua.edu.cn)

HAO ZHANG Department of Electronic Engineering Tsinghua University Beijing, 100084, P.R. China (e-mail: haozhang@tsinghua.edu.cn)

**Abstract:** A wealth of angle problems occur when facial recognition is performed: At present, the feature extraction network presents eigenvectors with large differences between the frontal face and profile face recognition of the same person in many cases. For this reason, the state-of-the-art facial recognition network will use multiple samples for the same target to ensure that eigenvector differences caused by angles are ignored during training. However, there is another solution available, which is to generate frontal face images with profile face images before recognition. In this paper, we proposed a method of generating frontal faces with image-to-image profile faces based on Generative Adversarial Network (GAN).

**Introduction**

Exceptional success has been achieved in facial recognition with the development of deep learning. However, many contemporary facial recognition models still have relatively poor performance in processing profile faces compared to frontal faces. Two methods have been proposed to solve this problem. One is to collect users' face images from all directions and train the classifier with both profile faces and frontal faces. Apple's FaceID is a typical example which requires users to turn their faces during registration. However, in more application scenarios, only frontal face images are available. By reproducing the classic face recognition network, we found out that many recognition errors were caused by matching profile faces with others' frontal faces. The other method is to generate frontal faces using profile faces before facial recognition. A typical idea is 3D-reconstruction [1], which identifies the projection angles by marking the feature points in the 2D photo and then fits a 3D general face model to generate a 2D frontal face through 2D orthographic projection, and used to be a popular method for generating the frontal face from a profile face images. However, this method largely relies on the accuracy of the predicted projection angle, which is not easy to acquire. At the same time, due to the limitation of the general 3D model used, the generated 2D face cannot properly represent all the features of the original human face. With the development of GAN, the application of GAN to generate human faces has become a mature technology. However, the classic GAN [2] can only generate random faces from random noise vectors. A discriminator's judgment standard focuses on whether the face is realistic. For the generation of frontal faces from profile images, we also required that the generated frontal face and the profile face be from the same person. Previous works including StackGAN [3] make it possible to generate photos from described characteristics. We can also obtain characteristics from the profile face and use them to generate a frontal face. To this end, we proposed an encoder & decoder end-to-end generator architecture.

Like many facial recognition networks, the encoder extracts characteristics of the profile face images into eigenvectors by using convolution and pooling, and then the decoder uses deconvolution to reconstruct the original face. This process is quite similar to what happens when an informant describes the appearance of a suspect to the police. The police does the suspect's portrait accordingly. In previous works, great breakthroughs have been made in both the encoder and the decoder. In the area of facial recognition, trained with triplet loss, FaceNet [4] has obtained accuracy higher than humans. Generator GAN like DCGAN [5] can also generate high-quality images. This means, in the aforementioned story, we have an informant with good language skills and a police officer with excellent drawing techniques. However, we must ensure that the informant and the police officer speak the same language and they understand each other well. This is why we cannot directly use both pre-trained FaceNet's feature extraction network and pre-trained DCGAN's generator. For this reason, most previous research on the encoder & the decoder GAN, such as Cycle-GAN [6] and Style-GAN [7], choose to train both the encoder and the decoder at the same time. However, to guarantee the quality of generated images, the decoder tends to have many activation functions with a limited output range, resulting in the vanishing gradient problem when back propagation is carried out. To solve this problem, a new separate training method has been introduced: We first pre-trained an encoder with an extra facial recognition task. It is noteworthy that triplet loss was not used in our work, because this would ignore some common characteristics shared by most individuals, which can be essential for future generation. With a fixed pre-training encoder, we used DCGAN's generator as a decoder. We trained the decoder with random weight initialization to ensure that the decoder can learn to understand the features extracted by the encoder during the training process.

As is mentioned earlier, the task of generating a frontal face from profile face images should guarantee not only that the generated frontal face image is realistic, but also that the input profile face images are from the same person. Classic GAN discriminators cannot evaluate these two different aspects at the same time. Authors of DR-GAN [8] added fake as a parallel label as identification. Thus, they can use one discrimator network to distinguish right from wrong and classify the identity at the same time. However, much work has been done on cheating the facial recognition network using a picture of simple colorful lines. These pictures can give the feature extraction network the same feature as a human face although they do not even look like one. In DR-GAN's practice, those generated pictures may be identified successfully although they look like faces. A triple GAN architecture has been proposed in [9], which divided the discriminator into two parts: a classifier and a discriminator to solve the problem of taxonomy generation. This ensured that both the classifier and the generator could achieve their own optimum respectively from the perspective of game theory and enabled the generator to sample data in a specific class. It's very similar to our task, except that we did not have a specific identity to generate. Inspired by the triple GAN architecture, we proposed a new GAN architecture consisted of a generator and two discriminators: one used for evaluating the authenticity of the generated frontal face and the other for judging whether the generated frontal face is from the same person as the profile face. The generator loss is the sum of losses from these two parts.

(Structure of the GAN system)

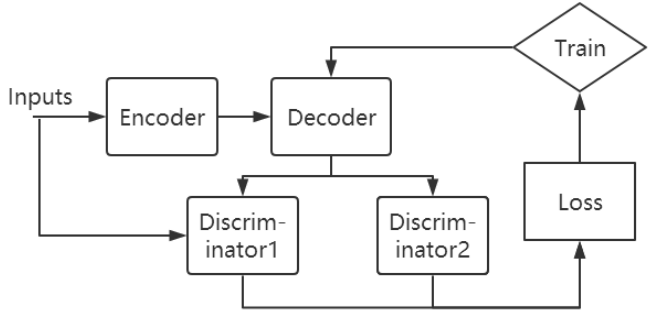

Increasing the number of discriminators may greatly unbalance the original Siamese structure of GAN, and asynchronizes the convergence rate between the generator and the discriminators, which will cause serious problems. Because each discriminator is trained independently, the discriminator convergence rate will not slow down. At the same time, since the frontal face generated by the generator must meet the requirements of both discriminators, there are two feasible directions to minimize the loss, which results in slower convergence. If we use the classic GAN training method, the discriminators will become more capable than the generator at the same training step. When training the GAN, usually we need a generator but we cannot directly use mature discriminators as the competitor because the discriminators' high accuracy will cause the vanishing gradient problem during the generator training, which has been proven in [10]. So we adopted wloss instead of the traditional cross-entropy loss, which makes loss to some extent indicate the training process, causing artificial control of the training process to be possible. We used artificial training control to largely avoid the capability imbalance between the generator and the discriminators and therefore improve the training efficiency. It will be explained in detail in the next section.

TP-GAN [11] used a two pathways to generate the frontal face: a local way for parts like eyes and noses and a global way for a rough whole face. They thought directly using the global way could not retain the characteristics of the original profile face properly. In our opinion, this was caused by the model collapse problem, which means in a given training set, only part of the patterns can appear in the generated results. This is fatal for generating frontal faces from the profile face images: For different individuals' input profile faces, the generator may obtain the same frontal face. Although the discriminator can increase the loss when model collapse happens, making the generator change, in many cases, this can just shift the generator distribution from one wrong pattern to another. In PacGAN [12], the author modified the discriminators so that they could judge the authenticity of more than one generated images simultaneously and the loss was relayed among all the generated images. Therefore, the loss will have a relatively much sharper rise when model collapse happens. In their case, the number of patterns appeared in the training color-MNIST dataset was increased from 300/1000 to 900/1000. In our work, the structure of PacGAN was introduced to our discriminator to judge the generated image's authenticity. This enabled our generator to gain face images with various and clear characteristics.

In conclusion, our contributions in this work mainly focus on the following:
1. Proposed a dual-discriminator structure of GAN to generate frontal faces from profile faces.
2. Fixed a pre-trained encoder during GAN training and proposed a controlled GAN training strategy to improve training efficiency.
3. Adopted PacGAN's discriminator architecture in the frontal face generation area, to improve the diversity of the generated patterns, making future facial recognition possible.

**Approach and Model Architecture**

Architecture（小标题）

For generators, we used the encoder-decoder end-to-end architecture: the encoder converted the input image into eigenvectors; the decoder reconstructed the frontal face with eigenvectors.

(Structure of the generator)

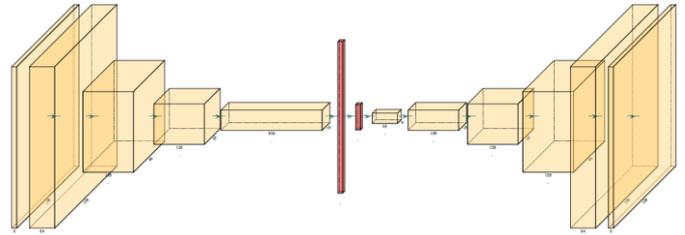

For the encoder, we used the traditional deep convolution structure to extract 512-dimensional eigenvectors from 128 x 128 x 3 original images.

Encoder

| Name | Input | Output | Kernel-size | Stride |
|---|---|---|---|---|
| Conv1 | 128,128,3 | 128,128,64 | 7 | 1 |
| Conv2 | 128,128,64 | 64,64,128 | 3 | 2 |
| Conv3 | 64,64,128 | 32,32,128 | 3 | 2 |
| Conv4 | 32,32,128 | 16,16,256 | 3 | 2 |
| Conv5 | 16,16,256 | 8,8,256 | 3 | 2 |
| Flatten | 8,8,256 | 16384 | | |
| Dense | 16384 | 512 | | |

For the decoder, we used the deconvolution structure of DCGAN to restore 128 x 128 x 3 frontal faces from 512-dimensional eigenvectors.

Decoder

| Name | Input | Output | Kernel-size | Stride |
|---|---|---|---|---|
| Dense | 512 | 8,8,64 | | |
| Conv1 | 8,8,64 | 8,8,64 | 8 | |
| Deconv1 | 8,8,64 | 16,16,128 | 3 | |
| Deconv2 | 16,16,128 | 32,32,128 | 3 | |
| Deconv3 | 32,32,128 | 64,64,128 | 3 | |
| Deconv4 | 64,64,128 | 128,128,64 | 3 | |
| Conv2 | 128,128,64 | 128,128,3 | 3 | |

For discriminator 1, which was used to decide whether the generated frontal face and original profile face belonged to the same person, we used the composition of the two images as the input.

Discriminator 1

| Name | Input | Output | Kernel-size | Stride |
|---|---|---|---|---|
| Conv1 | 128,128,6 | 64,64,128 | 3 | 2 |
| Conv2 | 64,64,128 | 32,32,128 | 3 | 2 |

| Name | Input | Output | Kernel-size | Stride |
|---|---|---|---|---|
| Conv3 | 32,32,128 | 16,16,256 | 3 | 2 |
| Conv4 | 16,16,256 | 8,8,256 | 3 | 2 |
| Conv5 | 8,8,256 | 4,4,256 | 3 | 2 |
| Flatten | 4,4,256 | 4096 | | |
| Dense | 4096 | 1 | | |

For discriminator 2, which was used to decide whether the input image was a real face or not, we adopted PacGAN's 4-image structure. We stitched 4 images into a double-size one as the input of the discriminator.

Discriminator 2

| Name | Input | Output | Kernel-size | Stride |
|---|---|---|---|---|
| Conv1 | 265,256,3 | 128,128,64 | 3 | 2 |
| Conv2 | 128,128,64 | 64,64,128 | 3 | 2 |
| Conv3 | 64,64,128 | 32,32,128 | 3 | 2 |
| Conv4 | 32,32,128 | 16,16,256 | 3 | 2 |
| Conv5 | 16,16,256 | 8,8,256 | 3 | 2 |
| Conv6 | 8,8,256 | 4,4,256 | 3 | 2 |
| Flatten | 4,4,256 | 4096 | | |
| Dense | 4096 | 1 | | |

Training Process （小标题）

For the discriminators, two independent discriminator networks were applied to judge the authenticity of the images and determine whether the generated images and the original images are the same person. For the first part, we used 4 generated image mosaics as the input. The discriminators' loss is the sum of the two discriminators' losses.

For the generator, we used an encoder-decoder structure, where the encoder was pre-trained with a facial recognition task. The generator loss was calculated by adding the losses of discriminator 1 and discriminator 2. To optimize the generator, competitive abilities against discriminator 1 and discriminator 2 needed to be optimized at the same time.

Although the fixed encoder could preserve the original Siamese network structure of GAN as much as possible, the multiple feasible directions made the generator still converge much slower than the discriminators.

In the design of the original GAN, if the generator's convergence rate is significantly slower than the discriminators', potentially, a mature discriminator may appear while the performance of the generator is still less satisfying. This makes it much harder for the generator to find a possible feasible direction, because no matter what tiny optimization the generator makes, the discriminator can still confidently distinguish true from false. This may eventually cause the vanishing gradient problem, which has been proven in [9]. To solve this problem, they proposed a simple correction method: By setting a maximum value for the convergence gradient, they roughly balanced the training process of the generator and the discriminator.

In our case, the generator is much harder to train than both of the discriminators. To solve this problem, we designed a GAN training process with artificial intervention to ensure that the discriminators will not be overtrained:

If discriminator 1_loss>-0.8 or if discriminator 2_loss>-0.8:
    If discriminator 1_loss > discriminator 2_loss
        Train discriminator 1
    Else
        Train discriminator 2
Else
    If Generator_loss_1>-0.8 and Generator_loss_2>-0.8
        Change loss_weight:
        (1* Generator_loss_1+1* Generator_loss_2)
        Train Generator
    Else
        If Generator_loss_1> Generator_loss_2
            Change loss_weight:
            (1* Generator_loss_1+0* Generator_loss_2)
            Train Generator
        Else
            Change loss_weight:
            (1* Generator_loss_1+0* Generator_loss_2)
            Train Generator
If Generator_loss_1<-0.8 and Generator_loss_2<-0.8
    Train discriminator_1
    Train_discriminator_2

**Experiment**

Training（小标题）

The Bosphorus dataset included 2D & 3D samples with clear shooting angle labels. We only used 2D data, including 4,666 pictures of 105 individuals.
(Samples in the Bosphorus dataset)

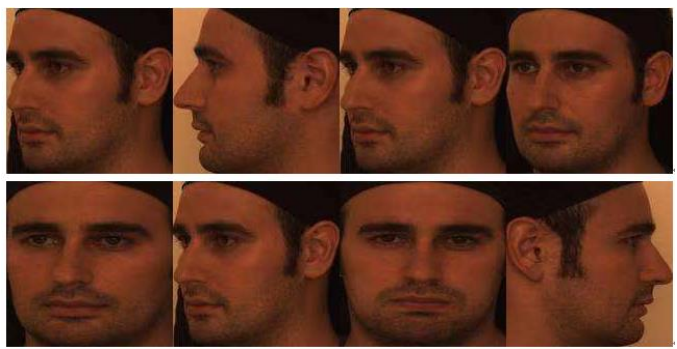

To form a training set, profile face images were chosen for input, and frontal face images for Ground Truth.
(Batch input profile faces)

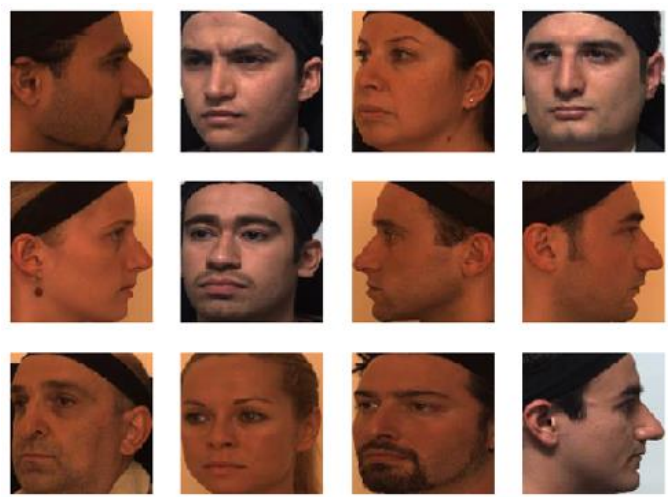

(Batch output frontal faces)

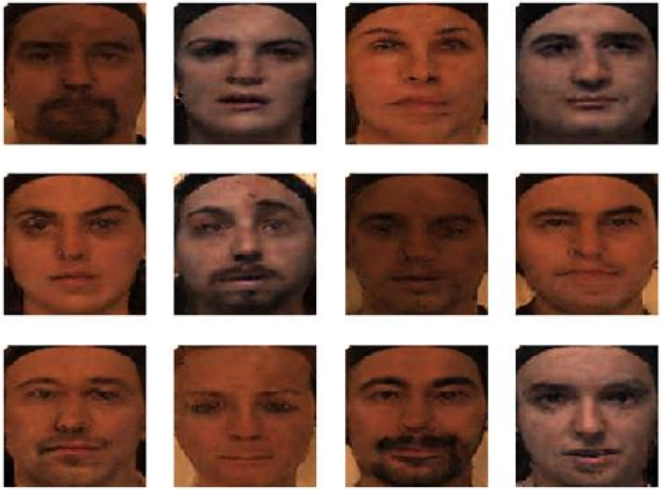

For contrast, with the same network structure, we used the classic GAN training method, which trains the generator and discriminators in turn.

(Comparative experiment output)

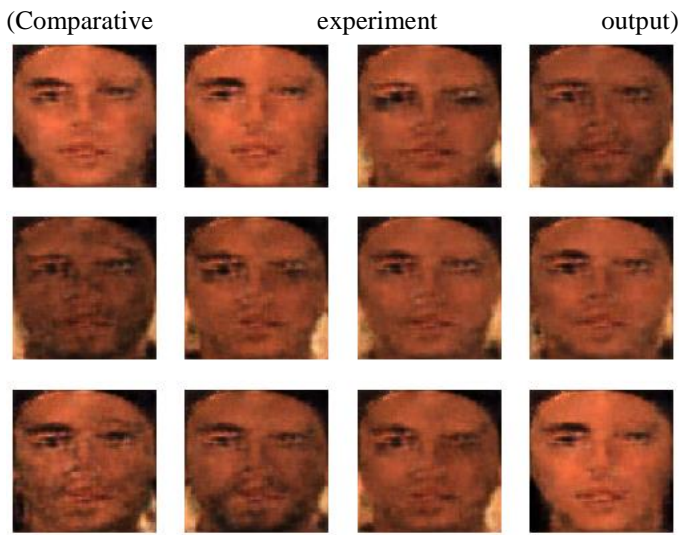

It can be concluded that because it was more difficult to train the generator than the discriminators and no training intervention was made, the discriminators get enough training far ahead of the generator. Meanwhile, the relatively strong discriminators made most of the generator movement useless, stopping it from finding a right feasible direction.

Test（小标题）

It can be seen that the generated images retained most of the facial features of the input images after end-to-end conversion from profile faces to frontal face. Because the generator loss was the total of the two discriminators' losses, the authenticity loss was not as important as it used to be. The authenticity of the image may not be as good as the classical generated random images. However, it is more important to retain features for facial recognition in the future. We think the tiny loss on authenticity of the image is acceptable.

Theoretically, it was difficult for the faces generated by GAN to fool state-of-the-art facial recognition networks like FaceNet, because during the training process, the generator's opponent discriminator had fewer layers and a less advanced architecture than professional facial recognition networks. Additionally, the performance of the generator and the discriminators was improved by turns during the training process, which meant, the competitive ability of the generator was about the same as that of the immature simple facial recognition network. However, in the practice of generating a frontal face with profile face images, the generator can learn enough facial features with a properly pre-trained encoder. These features help the generator to form a frontal face that retains as many original face features as possible.

We used FaceNet to build a classifier trained with individuals' real frontal faces, and let the classifier categorize the generated frontal face images. Among these generated images, 16.4% of the samples successfully fool the classifier into believing that they were true frontal face images of the input person.

Future Work

Future studies will focus on using more complex discriminator and generator models to improve output image quality. Although the current output images can be considered to have completed generation of frontal face from profile faces through examination with naked eyes, the image generated by GAN still has some unnatural characteristics which can be detected by state-of-the-art feature extraction networks. We will try to find a better generation network to solve this problem.

Conclusion

In this paper, we improved the method of training the end-to-end encoder & decoder generator network by pre-training the encoder separately on a related supervised learning task and proposed a training intervention strategy. A generator and dual-discriminator structure was proposed to meet the requirements of both authenticity and similarity in the facial recognition task. In addition, PacGAN was introduced to the frontal face generating area to deal with model collapse. Experiments showed that the frontal face generated with the profile face images to some extent can fool state-of-the-art facial recognition networks. If better image quality can be achieved by using a more complex decoder, GAN-assisted frontal face generation will be able to improve the accuracy of profile face recognition, which will be a great breakthrough in both facial recognition and GAN application.